\documentclass[conference]{IEEEtran}
\IEEEoverridecommandlockouts
\usepackage{cite}
\usepackage{amsmath,amssymb,amsfonts}
\usepackage{algorithmic}
\usepackage{graphicx}
\usepackage{textcomp}
\usepackage{xcolor}
\usepackage{makecell}
\usepackage{hyperref}

\def\BibTeX{{\rm B\kern-.05em{\sc i\kern-.025em b}\kern-.08em
    T\kern-.1667em\lower.7ex\hbox{E}\kern-.125emX}}
\begin{document}

\title{\LARGE Enhancing Speech Emotion Recognition Leveraging Aligning Timestamps of ASR Transcripts and Speaker Diarization\\
}

\author{
\begin{minipage}[t]{0.31\textwidth}
    \centering
    \fontsize{10pt}{12pt}\selectfont
    Hsuan-Yu Wang\\
    \textit{Department of English} \\
    \textit{National Taiwan Normal University} \\
    Taipei, Taiwan \\
    40921126l@ntnu.edu.tw
\end{minipage}
\hfill
\begin{minipage}[t]{0.31\textwidth}
    \centering
    \fontsize{10pt}{12pt}\selectfont
    Pei-Ying Lee\\
    \textit{Department of Computer Science and Information Engineering} \\
    \textit{National Taiwan Normal University} \\
    Taipei, Taiwan \\
    60947089S@ntnu.edu.tw
\end{minipage}
\hfill
\begin{minipage}[t]{0.31\textwidth}
    \centering
    \fontsize{10pt}{12pt}\selectfont
    Berlin Chen\\
    \textit{Department of Computer Science and Information Engineering} \\
    \textit{National Taiwan Normal University} \\
    Taipei, Taiwan \\
    berlin@ntnu.edu.tw
\end{minipage}
}

\maketitle

\begin{abstract}
In this paper, we investigate the impact of incorporating timestamp-based alignment between Automatic Speech Recognition (ASR) transcripts and Speaker Diarization (SD) outputs on Speech Emotion Recognition (SER) accuracy. Misalignment between these two modalities often reduces the reliability of multimodal emotion recognition systems, particularly in conversational contexts. To address this issue, we introduce an alignment pipeline utilizing pre-trained ASR and speaker diarization models, systematically synchronizing timestamps to generate accurately labeled speaker segments. Our multimodal approach combines textual embeddings extracted via RoBERTa with audio embeddings from Wav2Vec, leveraging cross-attention fusion enhanced by a gating mechanism. Experimental evaluations on the IEMOCAP benchmark dataset demonstrate that precise timestamp alignment improves SER accuracy, outperforming baseline methods that lack synchronization. The results highlight the critical importance of temporal alignment, demonstrating its effectiveness in enhancing overall emotion recognition accuracy and providing a foundation for robust multimodal emotion analysis.
\end{abstract}

\begin{IEEEkeywords}
Speech Emotion Recognition, Automatic Speech Recognition, Speaker Diarization
\end{IEEEkeywords}

\section{Introduction}
Speech Emotion Recognition (SER) has gained substantial research attention, particularly for its applications in human-computer interaction. While significant advancements have been made, real-world conversational scenarios pose unique challenges. Traditional SER systems often rely on manually segmented utterances, which are impractical to obtain at scale. Consequently, recent efforts have focused on integrating components like Speaker Diarization (SD) and Automatic Speech Recognition (ASR) to enable more autonomous emotion analysis from raw audio \cite{wu2023interspeech}. However, a fundamental limitation persists within these integrated approaches: the lack of precise temporal synchronization between ASR transcripts and speaker diarization outputs. This inherent misalignment can severely compromise the reliability of multimodal emotion recognition systems, especially in dynamic, turn-taking dialogues.

To overcome this critical challenge, we introduce a timestamp alignment pipeline that systematically matches ASR transcripts with speaker diarization segments. This process ensures that each transcribed sentence is accurately matched to its speaker and temporal boundaries, yielding high-quality, speaker-attributed utterances. Using this aligned data, our multimodal SER system combines RoBERTa-based textual embeddings \cite{liu2019roberta} with wav2vec 2.0-derived acoustic features \cite{baevski2020wav2vec}, wherein a cross-attention fusion mechanism, augmented by a gating component, integrates these modalities effectively \cite{jiang2022gated}. Evaluations on the IEMOCAP dataset \cite{2008IEMOCAP}demonstrate that precise alignment significantly improves SER accuracy, outperforming baseline systems lacking synchronization. These results emphasize the essential role of alignment in enhancing the robustness and reliability of multimodal emotion recognition.

\section{Related Work}
A recent line of research has addressed the limitations of manual segmentation in Speech Emotion Recognition (SER) by integrating Automatic Speech Recognition (ASR), speaker diarization (SD), and Voice Activity Detection (VAD) into fully automatic pipelines. While such integration improves scalability, it introduces challenges—particularly due to temporal misalignment between outputs from these components.

Whisper \cite{bain2023whisperx}, a self-supervised ASR model with fine-grained timestamping, has facilitated more precise speaker attribution and segment alignment. Building on this, \cite{li2024slt} demonstrated that even minor misalignments between ASR transcripts and speaker turns can significantly degrade SER performance. These works underscore the importance of synchronization but fall short of providing flexible, modular approaches that split alignment from end-to-end learning.

Meanwhile, a number of works have improved SER through architectural and learning-based innovations. \cite{rajapakshe2024emodarts} introduced a NAS-based framework to jointly optimize CNN and sequential modules, while  \cite{lu2020asr} and \cite{li2025centerloss} showed that pre-trained ASR features and compact representation learning respectively enhance emotion classification. \cite{wu2024multimodal} advanced modality fusion via a multi-loss framework, and \cite{zhao2019dyadic} incorporated conversational context in dyadic settings. \cite{xia2021temporal} emphasized the need to model temporal dynamics in emotion progression. These efforts contribute to downstream accuracy. However, these works often assume well-aligned inputs and overlook the critical preprocessing stage. Our approach addresses this upstream alignment challenge—an implicit dependency in many SER models.

\begin{figure*}{}
\centering
\includegraphics[width=1.0\textwidth]{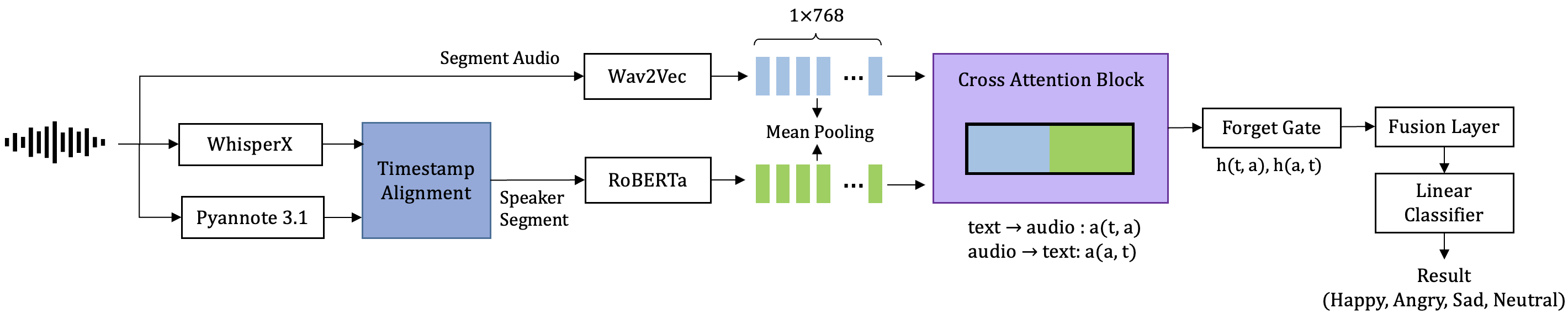}
\caption{Proposed Model}
\end{figure*}

Specific attention has also been paid to joint modeling of ASR and SD.  \cite{kanda2022transcribe} proposed an end-to-end speaker-attributed ASR model, while another approach \cite{kanda2022sa} uses encoder-decoder attractors to jointly infer transcripts and speaker turns. \cite{wang2023sed} formally introduced the Speech Emotion Diarization (SED) task and proposed the EDER metric to assess time-sensitive accuracy. \cite{neumann2024meeting} demonstrated that syntactic alignment cues further improve speaker attribution in realistic meetings. While effective, these methods are often tightly coupled or application-specific. In contrast, our modular alignment pipeline can be flexibly integrated with pre-trained ASR and SD models to enable more robust and transferable multimodal SER.

\section{Proposed Model}
This section sheds light on our proposed alignment-based multimodal SER framework, designed to robustly recognize speech emotions in conversational contexts. Our framework is structured around two core sections: a precise timestamp-based alignment pipeline for ASR and SD outputs, and a sophisticated cross-attention fusion mechanism for multimodal embeddings. The overall architecture is schematically depicted in Figure 1.
\subsection {ASR, SD, and Alignment Approach}
Accurate temporal synchronization between spoken words and their corresponding speakers is essential for reliable multimodal emotion recognition in conversational contexts. Our system begins by processing raw input audio through two independent, state-of-the-art models: WhisperX for ASR \cite{bain2023whisperx} and Pyannote 3.1 for SD \cite{bredin2020pyannote}. WhisperX generates detailed word-level transcripts with precise timestamps (e.g., [6.92s → 7.16s] "Excuse", [7.16s → 7.23s] "me"), while Pyannote identifies speaker turns by producing timestamped speaker-labeled segments (e.g., [6.92s → 7.24s] Speaker\_00). These two outputs serve as the foundation for our alignment procedure, which systematically links the transcribed text to the appropriate speaker segments. The entire process is visualized in Figure 2.

\begin{figure}[htbp]
\centerline{\includegraphics[width=0.5\textwidth]{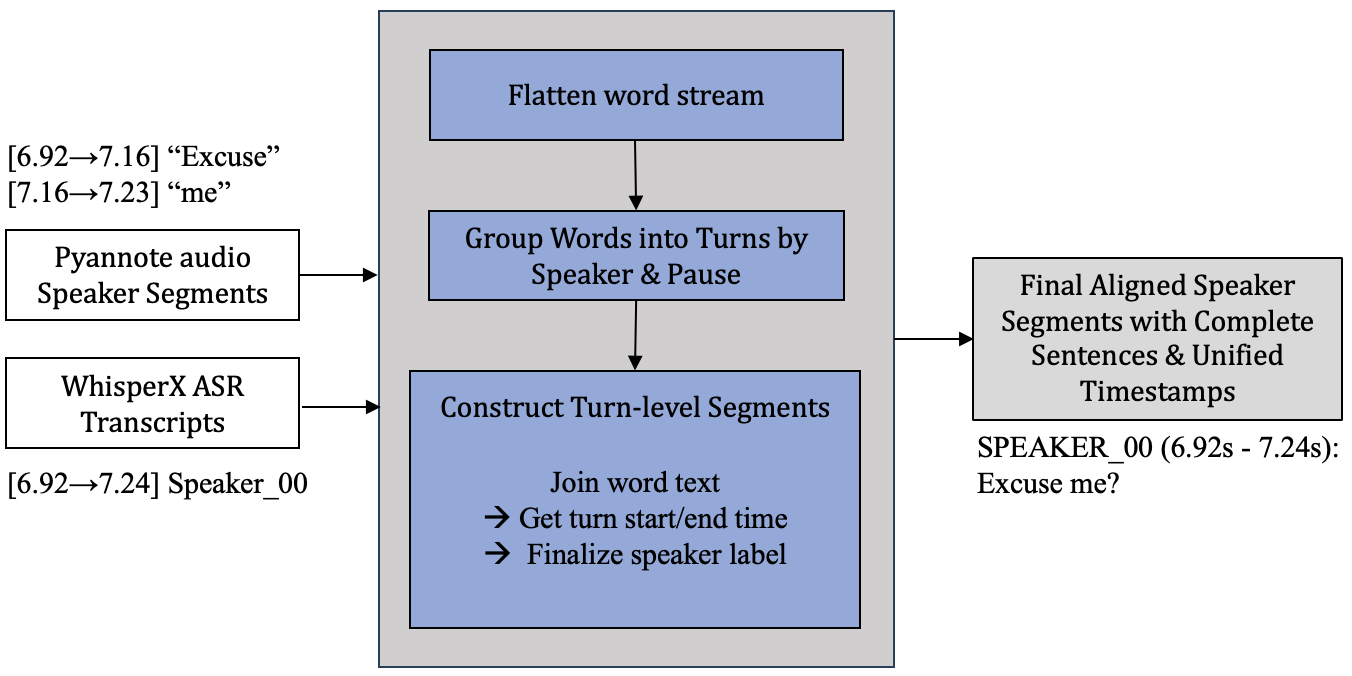}}
\caption{Timestamp Alignment Block}
\label{fig}
\end{figure}

\par To overcome the fragility of sentence-level text matching and mitigate data loss from ASR errors, we developed a robust, time- and speaker-aware alignment pipeline. This process, illustrated in Figure 2, transforms the word-level ASR and diarization outputs into coherent conversational turns. The procedure unfolds in three main stages:

\begin{itemize}
\item \textbf{Flatten Word Stream:} The process begins by integrating the output from the upstream ASR and diarization stage. All speaker-attributed words, complete with their individual timestamps, are extracted from their initial segments and organized into a single, chronologically ordered stream. This step effectively dissolves the preliminary, and often fragmented, ASR segment boundaries to create a continuous flow of word-level data.

\item \textbf{Group Words into Turns by Speaker \& Pause:} Next, this flat stream of words is grouped into semantically and contextually coherent turns. The grouping algorithm iterates through the words, merging consecutive words into a single turn as long as two conditions are met: (1) the speaker label remains consistent, and (2) the temporal pause between consecutive words does not exceed a predefined threshold (e.g., 1.5 seconds). A new turn starts whenever a speaker change or a significant pause is detected. This ensures that natural conversational breaks are respected while maintaining contextual continuity.

\item \textbf{Construct Turn-level Segments:} Finally, each grouped turn is finalized into a single, unified segment. The full text of the segment is constructed by joining the constituent words. Its overall timestamp is defined by the start time of the first word and the end time of the last word in the turn. The consistent speaker label is then assigned to this newly constructed segment.
\end{itemize}

\par This methodology ensures that the final speaker-attributed segments are resilient to minor ASR transcription errors and accurately reflect the natural flow of a conversation, providing high-quality, contextually rich inputs for the downstream emotion recognition model.

\begin{figure}[htbp]
\centerline{\includegraphics[width=0.5\textwidth]{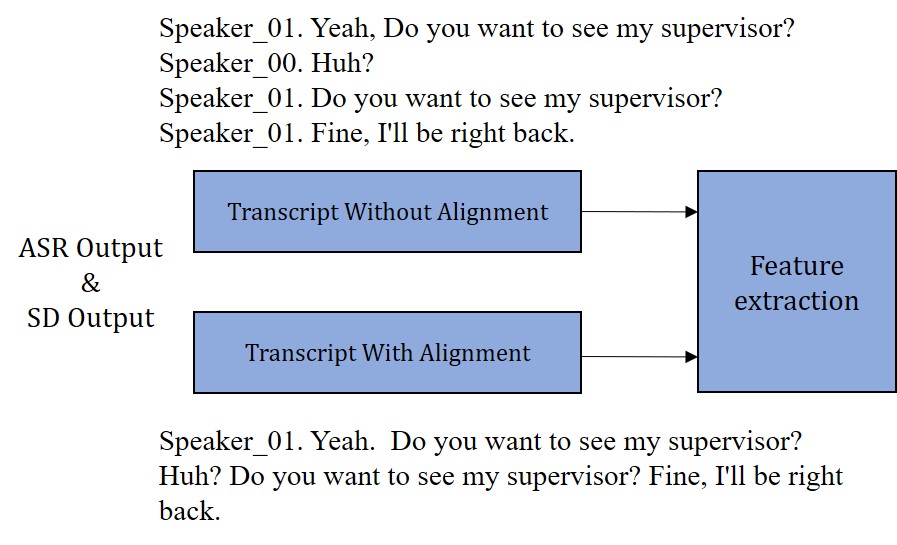}}
\caption{Transcript Comparison: without vs. with Alignment}
\label{fig}
\end{figure}

Figure 3 demonstrates the impact of our timestamp alignment. The 'Transcript Without Alignment' section illustrates common challenges in systems lacking precise temporal synchronization: fragmented ASR output and potential wrong speaker identification. As seen, the sentence is broken into very short, discrete segments, often leading to incomplete utterances. This fragmentation risks losing crucial contextual information, as the full emotion conveyed might only be apparent when considering the entire thought or conversational turn. Furthermore, short utterances can be mislabeled, as exemplified by a segment incorrectly attributed to another speaker, further corrupting input for emotion analysis.

In contrast, the 'Transcript With Alignment,' generated by our proposed alignment block, showcases significant improvements. Our process merges these shorter, fragmented ASR outputs based on temporal proximity and speaker consistency, resulting in longer, coherent conversational turns. This extended context allows the model to better capture the temporal evolution of emotions. Importantly, the alignment also corrects speaker misattributions and ensures that each segment is a complete utterance from a single, correctly identified speaker, providing cleaner, more reliable input for subsequent feature extraction.

\subsection{Bimodal Cross-Attention Fusion and Linear Classifier}
Following the precise timestamp alignment, the system proceeds to extract multimodal embeddings from the refined speaker-attributed segments, forming the core of our emotion recognition pipeline, as depicted in Figure 1. For each aligned segment, its audio waveform is processed by a wav2vec 2.0 model to generate rich, contextualized audio embeddings \cite{baevski2020wav2vec}. The output of  wav2vec 2.0, a sequence of audio features $\mathbb{R}^{T \times 768}$, undergoes mean pooling across the time dimension to yield a fixed-size segment-level audio embedding of $\mathbb{R}^{1 \times 768}$ \cite{reimers2019sbert}. 
\par Meanwhile, the corresponding aligned textual content is fed into a RoBERTa model, which produces contextualized text embeddings \cite{liu2019roberta}. Similar to wav2vec 2.0, RoBERTa's output sequence $\mathbb{R}^{T' \times 768}$ is mean-pooled across the token dimension to derive a segment-level text embedding of $\mathbb{R}^{1 \times 768}$. These segment-level embeddings are then prepared for the subsequent cross-attention operation. Both audio and text embeddings are then unsqueezed to $\mathbb{R}^{1 \times 1 \times 768}$ to ensure compatibility for the subsequent cross-attention operation. These processed audio and text embeddings serve as the inputs to our multimodal fusion strategy, which aims to leverage the complementary strengths of both modalities for robust emotion recognition. The final stages involve a fusion layer and a linear classifier to predict the emotion.

\par The extracted audio and text embeddings are then integrated within a Cross-Attention Fusion Block to enable inter-modal information exchange. Our model explicitly adopts the cross-modality gated attention fusion approach proposed by \cite{jiang2022gated}. While their original work focused on fusing three modalities (text, audio, and video), we adapt this mechanism for our two-modality setup (text and audio). This block utilizes Multi-Head Attention (MHA) to allow each modality to selectively attend to relevant parts of the other:

\begin{itemize}
    \item \textbf{Text to Audio Attention:} The text embedding acts as the Query, while the audio embedding serves as the Key and Value. This operation generates a new representation $a(t,a)$, where textual features are enriched by acoustic context.
    \item \textbf{Audio to Text Attention:} Conversely, the audio embedding becomes the Query, with the text embedding as the Key and Value. This yields $a(a,t)$, a representation where acoustic features benefit from linguistic context.
    \item \textbf{Forget Gate:} Following cross-attention, a gated fusion mechanism is applied. For instance, the original text embedding $(z_t)$ and its corresponding audio-attended representation $(a(t,a))$ are fed into a forget gate. This gate dynamically controls the flow of information from the attended modality, fusing it with the original representation to produce a refined multimodal feature, $h(t,a)$. A symmetric process yields $h(a,t)$.
\end{itemize}

\par This mechanism helps focus on the most salient multimodal features while mitigating noise. The output of the forget gate is a gated representation of dimension $\mathbb{R}^{1 \times 2 \times 768}$, combining the refined multimodal features. Subsequently, this gated representation is fed into a Fusion Layer, implemented as a Transformer-based architecture. Inside this layer, which is implemented as a Transformer-based architecture, the features are first concatenated along the sequence dimension to form a combined representation of $\mathbb{R}^{1 \times 2 \times 768}$. The output is then mean-pooled across its second dimension to condense it into a single $\mathbb{R}^{1 \times 768}$ vector representing the final fused embedding. Finally, this fused embedding is passed to a linear classifier. This classifier maps the high-dimensional representation to the output space of emotion categories via a linear transformation (i.e., $\text{Dim}=768 \rightarrow \text{Classes}$), followed by a Softmax activation function to yield a probability distribution over emotion classes. The emotion with the highest probability (determined by $\arg\max$) is selected as the final prediction.

\section{Experimental Setup}
This section details the experimental methodology used to evaluate our proposed multimodal SER framework, including the dataset, various training strategies, and the evaluation metrics employed.

\subsection{Dataset}

Our experiments were conducted on the IEMOCAP dataset\cite{2008IEMOCAP}, a widely used benchmark for emotion recognition research, comprising approximately 12 hours of audio-visual recordings of dyadic conversations between actors in both improvised and scripted scenarios. A key feature of IEMOCAP is its rich annotation, with each conversational turn labeled with speaker ID and categorical emotion. For this study, we focused exclusively on the four primary emotion categories commonly recognized in the literature: \textit{happy}, \textit{sad}, \textit{angry}, and \textit{neutral}. This dataset provides a challenging yet representative environment for evaluating multimodal emotion recognition systems in conversational contexts due to its realistic interactions and the presence of multiple speakers.

\subsection{Training Strategies}

To thoroughly assess the impact of our proposed timestamp alignment and multimodal fusion approach, we designed several distinct training strategies:

\begin{itemize}
    \item Full Proposed Model (with Alignment):
This configuration represented our complete framework, where the input audio underwent the full timestamp alignment pipeline (as detailed in Section~3.1) to generate precisely attributed speaker segments. These aligned segments were then processed by the wav2vec 2.0 and RoBERTa feature extractors, followed by the cross-attention fusion block (with the forget gate), fusion layer, and finally the linear classifier. This setup was expected to demonstrate the benefits of accurate temporal synchronization.
    \item Model Without Alignment (Baseline):
To highlight the critical role of our alignment pipeline, we established a baseline model that omits this preprocessing step. In this configuration, ASR transcripts (e.g., from WhisperX) and raw speaker segments (from Pyannote) were used with a simpler, less precise method of association (e.g., overlap at the segment level without fine-grained word-level attribution or sentence reconstruction). The multimodal fusion architecture remained the same, allowing us to isolate the performance gains attributable solely to the timestamp alignment.
    \item Freezing Embedding Extractors:
For both the \textit{with alignment} and \textit{without alignment} setups, we conducted experiments where the pre-trained wav2vec 2.0 and RoBERTa embedding extractors were frozen during training. This strategy prevented fine-tuning of these large foundation models and primarily evaluated the capacity of the cross-attention fusion and subsequent layers to learn effective multimodal representations from fixed, high-quality embeddings. This facilitated to understand whether the performance gains were due to the fusion mechanism itself or also involved adaptive fine-tuning of the feature extractors.
\item Fine-tuning Embedding Extractors
Conversely, in another set of experiments, the wav2vec 2.0 and RoBERTa models were fine-tuned along with the rest of the network. This allowed the feature extractors to adapt their representations specifically for the SER task and the nuances of the IEMOCAP dataset, potentially leading to higher overall accuracy.
\end{itemize}

\subsection{Evaluation Metrics}

To rigorously evaluate the performance of our models, we employ standard classification metrics, as well as specialized metrics suited for SER in conversational contexts. Performance is primarily measured using:

\begin{itemize}
    \item \textbf{Weighted Average Recall (WAR / Accuracy):} Represents the overall classification accuracy across all emotion categories, weighted by the number of samples in each class.
\end{itemize}

Furthermore, acknowledging the complexities of evaluating emotion recognition on automatically segmented and diarized speech, we adopt the advanced metrics proposed by \cite{wu2023interspeech}:

\begin{itemize}
     \item \textbf{Time-weighted Emotion Error Rate (TEER):} A duration-aware metric that penalizes missed speech, false alarms, and emotion misclassifications. This provides a more ecologically valid measure of SER performance in continuous speech.

    \begin{equation}
    \mathrm{TEER} = \frac{\mathrm{MS} + \mathrm{FA} + \mathrm{CONF}_{\text{emo}}}{\mathrm{TOTAL}}
    \end{equation}

    \item \textbf{Speaker-Attributed TEER (sTEER):} Extends TEER by incorporating speaker attribution errors. It penalizes both incorrect emotion labels and incorrect speaker assignments, directly assessing the quality of automatic speaker-emotion alignment.

    \begin{equation}
    \mathrm{sTEER} = \frac{\mathrm{MS} + \mathrm{FA} + \mathrm{CONF}_{\text{emo+spk}}}{\mathrm{TOTAL}}
    \end{equation}
\end{itemize}

By utilizing these comprehensive metrics, we aim to provide a thorough and nuanced assessment of the effectiveness of our model, particularly in addressing the challenges of real-world, speaker-rich conversational SER.

\begin{table}[htbp]
\caption{Overall Results on Emotion Recognition}
\label{tab:overall_results}
\centering
\begin{tabular}{|l|c|c|c|}
\hline
\textbf{Setup} & \textbf{Accuracy (\%)} & \makecell[l]{\textbf{Weighted} \\ \textbf{F1-Score}} & \makecell[l]{\textbf{Macro} \\ \textbf{F1-Score}}\\
\hline
\makecell[l]{ Wu et.al \cite{wu2023interspeech} \\ (on 6 emotion categories)} & 49.49 & - & - \\
\hline
Without Alignment & 56.82 & 53.87 & 47.10\\
\hline
Proposed Approach & 66.81 & 66.88 & 66.48\\
\hline
\end{tabular}
\end{table}

\section{Experimental Results}
As shown in TABLE I, our proposed model, incorporating the timestamp alignment pipeline, achieves an overall accuracy of 66.81\%. This represents a notable improvement over the "Without Alignment" baseline, which achieves an accuracy of 56.82\%. This difference, while seemingly modest in raw accuracy, signifies the benefit of precise temporal synchronization. More importantly, the Weighted F1-score for our proposed model increases substantially from 53.87\% to 66.81\%, and the Macro F1-score sees a significant leap from 47.10\% to 66.48\%. The substantial improvement in Macro F1-score is particularly encouraging, as it indicates a much better balance in performance across all emotion categories, especially for minority classes.

\begin{table}[htbp]
\caption{F1-Score Comparison with and without Alignment}
\label{tab:f1_alignment_comparison}
\centering
\begin{tabular}{|l|c|c|}
\hline
\textbf{Emotion} & \makecell{\textbf{F1-score} \\ \textbf{(Without Alignment)}} & \makecell{\textbf{F1-score} \\ \textbf{(With Alignment)}}  \\
\hline
happy   & 0.64 & 0.73 \\
angry   & 0.38 & 0.64 \\
sad     & 0.26 & \textbf{0.67} \\
neutral & 0.61 & 0.62 \\
\hline
\end{tabular}
\end{table}

A key finding from our per-emotion analysis is the significant improvement in recognizing the "sad" category. As shown in TABLE II, the F1-score for "sadness" increases massively from 0.26 to 0.67 after implementing our alignment pipeline. This substantial gain highlights the critical role of temporal context for certain emotions.
This result aligns with existing research indicating that emotions like sadness are often characterized by subtle acoustic cues expressed over longer durations, requiring a broader temporal window for accurate recognition \cite{zhao2019dyadic} \cite{xia2021temporal}. Without proper alignment, conversational turns can be fragmented, breaking these essential long-term dependencies and causing the model to miss the defining features of sadness. Our alignment pipeline directly remedies this by reconstructing coherent, complete utterances. By providing the model with the full emotional expression, it is better equipped to capture the subtle and extended patterns characteristic of sadness, distinguishing it more effectively from other low-arousal states.

\begin{table}[htbp]
\caption{TEER and sTEER Comparison Across Pipelines}
\label{tab:teer_steer_summary}
\centering
\begin{tabular}{|l|c|c|}
\hline
\textbf{Pipeline} & \textbf{TEER (\%)\,↓} & \textbf{sTEER (\%)\,↓} \\
\hline
\makecell[l]{Wu et al.~\cite{wu2023interspeech} \\ (on 6 emotion categories)} & 66.03 & 65.17 \\
\hline
Standard Pipeline & 89.35 & 90.96 \\
\hline
VAD-Oracle Pipeline & 53.90 & 62.54 \\
\hline
\end{tabular}
\end{table}

To further assess the impact of accurate segmentation on overall system performance, we compare our standard pipeline with a VAD-Oracle configuration, in which perfect ground-truth segment boundaries are provided and the internal VAD is disabled. As shown in Table~\ref{tab:teer_steer_summary}, removing the influence of VAD yields a massive improvement: TEER drops from 89.35 to 53.9\%, and sTEER from 90.96 to 62.54\%. These results demonstrate that when segmentation quality is no longer a limiting factor, our alignment and speaker-attribution components are capable of operating near their optimal capacity. Compared to the baseline established in ~\cite{wu2023interspeech}, which reported TEER and sTEER of 66.03\% and 65.17\% respectively, our VAD-Oracle pipeline achieves lower errors across both metrics. This suggests that, while VAD remains a major bottleneck in practical systems, our alignment pipeline—when being free from segmentation noise—can outperform previously reported state-of-the-art results.

\begin{table}[htbp]
\caption{Performance Comparison of Embedding Extractor Strategies (With Alignment)}
\label{tab:finetuning_comparison}
\centering
\begin{tabular}{|l|c|c|c|}
\hline
\textbf{Strategy} & \textbf{Accuracy (\%)} & \textbf{Weighted F1 (\%)} & \textbf{Macro F1 (\%)} \\
\hline
\makecell[l]{Frozen \\ Embeddings} & 56.82 & 53.87 & 47.10 \\
\hline
\makecell[l]{Fine-tuned \\ Embeddings} & \textbf{66.81} & \textbf{66.81} & \textbf{66.48} \\
\hline
\end{tabular}
\end{table}

To further analyze the contribution of different components, we investigated the effect of fine-tuning the pre-trained wav2vec 2.0 and RoBERTa embedding extractors. As shown in Table IV, when the embedding extractors are frozen (i.e., their weights are kept fixed during training), our model achieves an accuracy of 56.82\%. In contrast, allowing the embedding extractors to be fine-tuned alongside the rest of the network leads to a notable increase in overall accuracy to 66.81\%. This improvement indicates that while the pre-trained wav2vec 2.0 and RoBERTa models provide strong general-purpose representations, enabling them to adapt their weights specifically to the nuances of the IEMOCAP dataset and the SER task significantly enhances overall performance. The ability of the fine-tuned extractors to learn more task-relevant features directly contributes to the improved classification accuracy. While specific F1-scores for the "frozen" setup are not detailed per emotion in the presented tables, the overall accuracy difference strongly suggests that fine-tuning also leads to better balanced performance across classes, implying improvements in both Weighted and Macro F1-scores akin to the gains observed with the alignment strategy. This highlights that optimized feature extraction, through fine-tuning, complements the benefits derived from our precise timestamp alignment.

\section{Conclusion and Future Work}
In this paper, we have put forward and validated a systematic timestamp alignment pipeline designed to resolve synchronization issues and reconstruct contextual coherence from the outputs of independent ASR and speaker diarization models for multimodal SER. Our core contribution is a multi-stage process that segments transcripts into meaningful sentences, attributes speakers at a sentence level, and merges them into conversational turns, thereby providing a high-quality, analysis-ready input for downstream tasks.

To demonstrate the efficacy of our alignment pipeline, we have evaluated its impact on a representative cross-attention fusion architecture. The results conclusively affirm our hypothesis: the model trained on data processed by our alignment pipeline significantly outperforms the baseline that used a simpler association method. This is most evident in the substantial increase of the Macro F1-score from 47.10\% to 66.48\%, driven by a huge improvement in recognizing context-dependent emotions like "sadness" (F1-score improved from 0.26 to 0.67).

This finding underscores that for complex conversational tasks like SER, meticulous front-end data processing is not merely a preliminary step but a fundamental determinant of model performance. Our work illustrates that the quality and contextual integrity of input segments are as crucial as the sophistication of the downstream fusion architecture. While this study focuses on IEMOCAP due to its detailed annotations and widespread adoption, future work will extend to multilingual and in-the-wild datasets to evaluate cross-domain generalization. We also plan to refine the turn re-segmentation logic, fine-tune the VAD component for improved boundary accuracy, and systematically assess the robustness of the proposed alignment pipeline across diverse multimodal fusion models.

\section*{Acknowledgment}

The authors would like to thank Jhih-Rong Guo, a member of National Taiwan Normal University's Speech and Machine Intelligence Laboratory (SMIL), for his valuable guidance and feedback throughout the development of this work.

\section*{Code Availability}
An open-source implementation of our method is available online.\footnote{\href{https://github.com/M4RC034/SER-Align.git}{https://github.com/M4RC034/SER-Align.git}}

\end{document}